\documentclass[11pt]{article}

\usepackage[preprint]{acl}

\usepackage{times}
\usepackage{latexsym}

\usepackage[T1]{fontenc}

\usepackage[utf8]{inputenc}

\usepackage{microtype}

\usepackage{inconsolata}

\usepackage{graphicx}

\usepackage{enumitem}
\usepackage{adjustbox}
\usepackage{amsmath}
\usepackage{amssymb}
\usepackage{amsthm}
\usepackage{cleveref}
\usepackage{stackengine}
\usepackage{comment}
\usepackage{mathtools}
\usepackage{xcolor}
\usepackage{longtable}
\usepackage{multirow}
\usepackage{graphicx}
\usepackage{subcaption}
\usepackage{colortbl}

\theoremstyle{plain}

\usepackage{xcolor}
\usepackage{cleveref}
\usepackage{enumitem}

\usepackage{algorithm}
\usepackage{algorithmicx}
\usepackage{algpseudocode}
\title{PlaM: Training-Free Plateau-Guided Model Merging\\for Better Visual Grounding in MLLMs}

\author{Zijing Wang$^{1}$ Yongkang Liu$^{1}$$^\dag$  Mingyang Wang$^{2,3}$ Ercong Nie$^{2,3}$ Deyuan Chen$^{1}$\\
  \textbf{Zhengjie Zhao$^{1}$ Shi Feng$^{1}$ Daling Wang$^{1}$$^\dag$ Xiaocui Yang$^{1}$ Yifei Zhang$^{1}$ and Hinrich Schütze$^{2,3}$} \\
        $^1$Northeastern University, China;
        $^2$CIS, LMU Munich, Germany \\
        $^3$Munich Center for Machine Learning (MCML), Germany \\
        \texttt{wzj1718@gmail.com}
}
\begin{document}
\maketitle

\def\thefootnote{$\dag$}\footnotetext{Corresponding Author}\def\thefootnote{\arabic{footnote}}
\begin{abstract}
Multimodal Large Language Models (MLLMs) rely on strong linguistic reasoning inherited from their base language models. However, multimodal instruction fine-tuning paradoxically degrades this text's reasoning capability, undermining multimodal performance. To address this issue, we propose a training-free framework to mitigate this degradation. Through layer-wise vision token masking, 
we reveal a common three-stage pattern in multimodal large language models: early-modal separation, mid-modal alignment, and late-modal degradation. By analyzing the behavior of MLLMs at different stages, we propose a plateau-guided model merging method that selectively injects base language model parameters into MLLMs. Experimental results based on five MLLMs on nine benchmarks demonstrate the effectiveness of our method. Attention-based analysis further reveals that merging shifts attention from diffuse, scattered patterns to focused localization on task-relevant visual regions.
Our repository is on \url{https://github.com/wzj1718/PlaM}.




\end{abstract}
\section{Introduction}

Multimodal Large Language Models (MLLMs) have attracted widespread attention. Representative systems such as GPT-4V~\cite{yang2023dawn}, Gemini~\cite{team2024gemini}, along with a growing ecosystem of open-source models (e.g., LLaVA-style~\cite{li2024llava,liu2024improvedbaselinesvisualinstruction} and Qwen-VL-style models~\cite{qwen2.5-VL,qwen3technicalreport}), demonstrate impressive multimodal instruction-following and visual reasoning capabilities. 

\begin{figure}[t]
    \centering
    \begin{minipage}[t]{0.32\columnwidth}
        \centering
        \includegraphics[width=\linewidth]{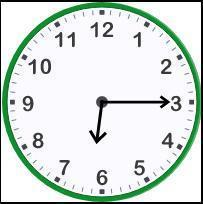}
    \end{minipage}\hfill
    \begin{minipage}[t]{0.32\columnwidth}
        \centering
        \includegraphics[width=\linewidth]{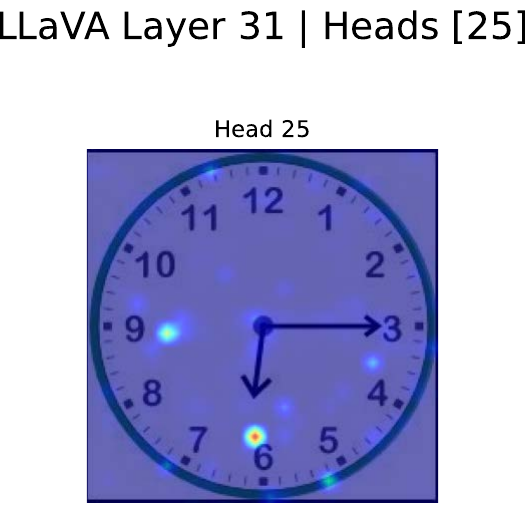}
    \end{minipage}\hfill
    \begin{minipage}[t]{0.32\columnwidth}
        \centering
        \includegraphics[width=\linewidth]{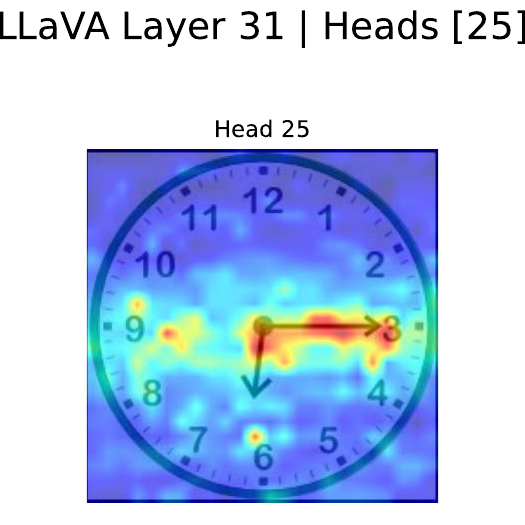}
    \end{minipage}
    \caption{Our proposed PlaM improves visual grounding and prediction. Given the clock image (left) and the question "It is (\_) past six.'', the original MLLM without PlaM attends diffusely and answers “half” (middle). After applying PlaM model merging, attention concentrates on the clock hands and the model correctly answers “quarter” (right).}
    \label{figure:intro-exam}
    \vspace{-1em}
\end{figure}


Most MLLMs are built by extending a strong text-only LLM and then applying multimodal instruction fine-tuning. Although this pipeline yields strong multimodal understanding and generation~\cite{alayrac2022flamingo}, mounting evidence suggests that the resulting capabilities remain largely text-dominant: models rely disproportionately on textual signals while under-utilizing visual evidence~\cite{wu2025language,zheng2025unveiling,zhao2025looking,li2025vista,chen2024image,liu2024paying}. This pattern appears across multiple architectures and tasks. For instance, an analysis of VideoLLaMA-7B shows that output tokens attend to text tokens 157 times more than to visual tokens~\cite{wu2025language}, and broader studies report that MLLMs often answer visual questions primarily using textual knowledge, with visual information playing only a secondary role~\cite{wu2025mitigating,liu2025modality}. Importantly, multimodal instruction fine-tuning can also degrade the base model’s original text reasoning ability
\cite{zhang2024wings,ratzlaff2025training,lu2024deepseek,li2024multi}. 
This degradation is particularly concerning because strong text reasoning is the backbone for composing and verifying multimodal inferences; when it weakens, models become less able to correctly interpret and leverage visual signals, further exacerbating poor visual grounding.
As illustrated in Figure~\ref{figure:intro-exam}, failure cases exhibit scattered and semantically irrelevant attention over the image, whereas correct predictions require focusing on task-relevant regions. Overall, multimodal instruction tuning weakens text processing and  fails to promote effective visual grounding, motivating recent efforts to address this trade-off~\cite{lu2024deepseek}.

Existing attempts to address language-reasoning degradation in MLLMs are largely training-based. Typical strategies include introducing auxiliary components to compensate attention shifts induced by visual token expansion~\cite{zhang2024wings}, replaying or interleaving text-only data during training~\cite{mckinzie2024mm1,li2024mini}, and using preference alignment objectives to jointly optimize visual instruction quality and language instruction-following~\cite{li2024multi}. However, these approaches often require additional computational resources, carefully curated data, or architectural modifications. In contrast, training-free solutions remain comparatively underexplored. \citet{ratzlaff2025training} show that the choice of the base language model and its inherent capability significantly influence the extent of text-task degradation after multimodal fine-tuning. Meanwhile, \citet{zhang2025evaluating} suggest that modality preference can be characterized by a direction in latent representations and steered via representation engineering, enabling control over modality preference without additional fine-tuning.



In this paper, we propose Plateau-guided model Merging (\textbf{PlaM}), a general and efficient training-free method to enhance the text reasoning capability of MLLMs. We begin with a layer-function analysis and show that the utilization of visual information in MLLMs follows a diminishing-returns pattern across decoder depth. Specifically, through layer-wise vision token masking, we reveal a consistent three-stage behavior: (1) in early layers, models struggle to effectively exploit visual evidence and exhibit insufficient visual grounding due to modal separation; (2) in middle layers, guided by textual context, MLLMs performs modal alignment and progressively attend to semantically critical visual features, leading to rapid performance gains; and (3) in late layers, modal alignment tends to stabilize, visual features are absorbed by textual information, and additional visual access yields little further improvement, resulting in a performance plateau. Importantly, we find that this late-layer plateau is closely linked to weakened language capability after multimodal tuning: degraded text reasoning limits the model's ability to guide attention toward the correct visual features, resulting in a performance bottleneck. To recover the compromised text ability, we introduce a plateau-guided model merging strategy that selectively injects base language model parameters into these underutilized late layers via linear interpolation. Extensive experiments across five representative MLLMs and nine benchmarks demonstrate consistent improvements, validating the effectiveness and generality of our training-free method.


Our contributions can be summarized as follows:
\begin{itemize}[noitemsep, left=0pt]
\item We identify a three-stage pattern of visual token utilization and demonstrate that late model layers exhibit diminishing returns with respect to additional visual access;
\item We propose plateau-guided model merging (PlaM), a simple training-free approach that restores base LM capabilities in late layers and achieves consistent improvements across multiple models and benchmarks;
\item We provide mechanistic insights through attention analysis, showing that performance gains arise from enhanced visual grounding that transforms diffuse attention into focused, task-relevant localization of visual evidence.
\end{itemize}

\section{Related Work}
\paragraph{MLLMs.} MLLMs have rapidly emerged as a promising paradigm for integrating language and vision within a unified system~\cite{yin2024survey,han2025surveygenerativecategoriestechniques}. Rather than processing each modality in isolation, these models are typically designed to support cross-modal interaction by mapping modality-specific inputs into a shared space that a language-centric reasoning core can operate on~\cite{radford2021learningtransferablevisualmodels,lu2019vilbert}. A growing body of work builds MLLMs by leveraging pretrained large language models as the primary reasoning engine, while attaching dedicated encoders (e.g., vision encoders) and lightweight alignment components to transform non-text signals into language-compatible representations~\cite{alayrac2022flamingo,li2023blip,tsimpoukelli2021multimodal,eichenberg2022magma,zhuminigpt}. Recent progress further shows that instruction-style multimodal fine-tuning, which often uses large-scale, automatically constructed or weakly supervised multimodal instruction data, can substantially improve multimodal instruction following and reasoning~\cite{liu2023visual,ye2023mplug,dai2023instructblip}. 
\paragraph{Mitigating Language Reasoning Degradation in MLLMs.} Multimodal instruction tuning often degrades a pretrained language model's original text-only reasoning capabilities. \textbf{Most existing solutions are training-based}:~\citet{wu2024controlmllm} manipulate attention responses via inference-time latent variable optimization to improve visual referring behavior.~\citet{zhang2024wings} introduce auxiliary architectural components to compensate for attention shifts induced by visual token expansion, while others preserve language reasoning by interleaving text-only data or using preference-based alignment~\cite{mckinzie2024mm1,li2024mini,li2024multi}.
However, these methods require additional training, curated datasets, or architectural modifications. \textbf{Training-free approaches remain relatively underexplored}.~\citet{ratzlaff2025training} show that text degradation depends heavily on base LLM choice,~\citet{wu2025language} compress non-text tokens to reshape attention allocation, and~\citet{zhang2025evaluating} steer modality preference via latent representation offsets, though this requires externally specifying modality preference. Our work proposes a training-free strategy that explicitly restores language model functionality in underutilized late layers, alleviating language reasoning degradation while improving visual grounding.



\section{Method}\label{lab:Preliminary}
In this section, we present \textbf{PlaM} (Plateau-guided model Merging), a training-free approach that improves visual grounding by restoring late-layer language reasoning in MLLMs. PlaM consists of two components used throughout the paper. First, we adopt a layer-wise vision token masking procedure to characterize how visual information is utilized across decoder depth and to identify where performance plateaus. Second, guided by this layer-wise characterization, we selectively merge parameters between the base language model and the fine-tuned MLLM within a chosen layer range, while keeping the vision encoder and projector fixed.

\paragraph{Notation} Given an image $I$ and a text prompt $T$, an MLLM encodes the two modalities into a unified token sequence. The image is first processed by a vision encoder $E_\text{vis}$, which maps the raw image into a set of high-dimensional visual features. These features are then projected into the language model's embedding space via a projector $P$, yielding visual tokens $
X^{(0)}_\text{vis}=P(E_\text{vis}(I)) \in \mathbb{R}^{N_\text{vis} \times  d}$.
The text prompt $T=(t_1,...,t_{N_\text{txt}})$ is embedded through the language model's token embedding layer, producing textual tokens $X^{(0)}_\text{txt} \in \mathbb{R}^{N_\text{txt} \times  d}$. The input to the language model is formed by concatenating visual and textual tokens: $X^{(0)}=[X^{(0)}_\text{vis};X^{(0)}_\text{txt}]\in \mathbb{R}^{N \times d}$, where $N=N_\text{vis}+N_\text{txt}$ is the total sequence length and $d$ is the hidden dimension of the language model.

The language model decoder consists of $L$ Transformer layers, and the hidden states evolve through the network as $X^{(l)}=\Phi ^{(l)}(X^{(l-1)}), \quad l=\{1,...,L \}$. During inference, the model first encodes the entire multimodal prompt to construct the KV cache, and then generates output tokens autoregressively conditioned on the cached representations.

\paragraph{Vision Token Masking} To quantify how MLLMs rely on visual information across decoder depth, we apply a depth-controlled masking strategy on visual tokens~\cite{shi2025vision}. Let $\mathcal{V} \subset \{N_1,N_1+1\dots,N_1+N_\text{vis}-1\}$ denote the index set of visual tokens in the input sequence. For a selected layer $k$, visual tokens are processed normally in all layers $l < k$. For layers $l \ge k$, we block access to visual tokens by removing their positions from the attention keys and values: $\mathrm{Attn}^{(l)}(Q,K,V) \;\Rightarrow\; \mathrm{Attn}^{(l)}\bigl(Q,\,K_{\neg \mathcal{V}},\,V_{\neg \mathcal{V}}\bigr)$. This intervention preserves the textual pathway while preventing later layers from attending to visual tokens. Sweeping $k$ from shallow to deep layers yields a layer-wise profile characterizing where, and to what extent, visual information contributes to the decoder’s predictions.

\paragraph{Model Merging Across Modalities}

We aim to enhance MLLMs' performance by selectively incorporating base language model parameters into specific layers of the fine-tuned model. Let $\mathbf{W}_{\text{vlm}}$ denote the parameters of the fine-tuned MLLM and $\mathbf{W}_{\text{lm}}$ denote those of the corresponding base language model (sharing the same backbone architecture). We first identify a target layer set $\mathcal{L}=\{l_m, l_{m+1}, \ldots, l_n\}$, typically in the middle-to-late decoder layers. For each layer $l \in \mathcal{L}$, we construct merged parameters: 
\[
\mathbf{W}_{\text{merged}}^{(l)} = \lambda_1 \mathbf{W}_{\text{lm}}^{(l)} + \lambda_2 \mathbf{W}_{\text{vlm}}^{(l)}, \quad \lambda_i \in [0,1.5].
\]
For layers outside $\mathcal{L}$, we preserve the original MLLM parameters: \[\mathbf{W}_{\text{merged}}^{(l)} = \mathbf{W}_{\text{vlm}}^{(l)}, \quad \forall l \notin \mathcal{L}.\]
The vision encoder and projector remain fixed, and only the selected backbone layers undergo merging. The hyperparameter $\lambda_i$ controls the degree to which we reintroduce the base language model's capabilities. By varying $\lambda_i$ and the layer range $\mathcal{L}$, we obtain a family of merged models that balance multimodal alignment with the general reasoning capabilities of the base model.

\begin{figure*}[t]
    \centering
    \includegraphics[width=1.0\linewidth]{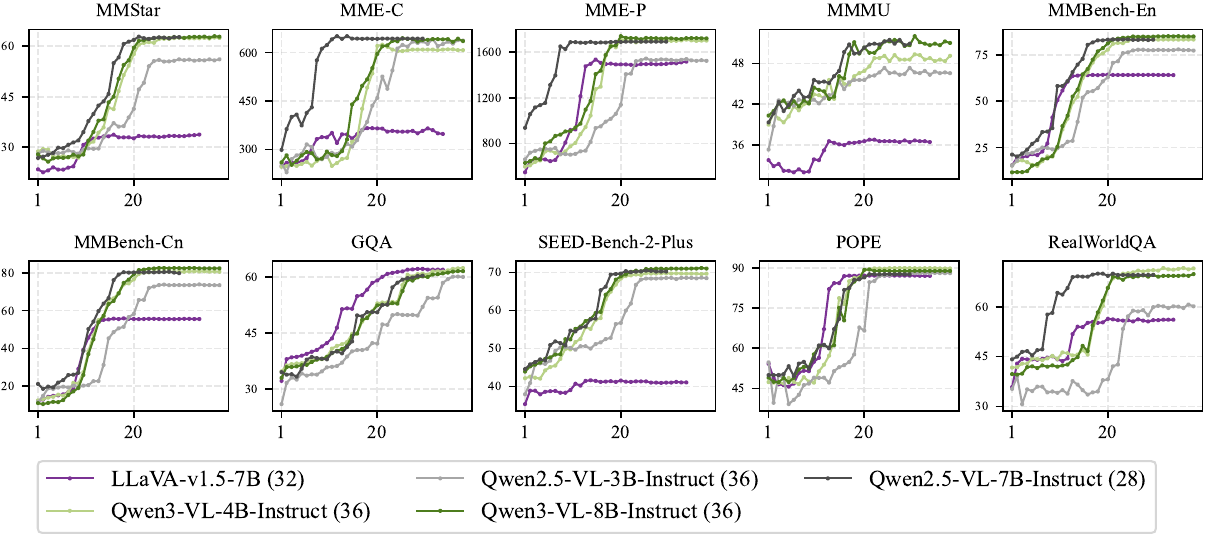}
    \caption{Performance versus cut layer ($k$) under depth-controlled vision token masking. Vision tokens are removed from layer $k$ onward ($k=L+1$ indicates no masking), and each curve reports the official metric on each benchmark. See~\Cref{tab:llavav1.5,tab:Qwen2.53B,tab:Qwen2.57B,tab:Qwen34B,tab:Qwen38B} for the corresponding numerical results (Appendix~\ref{app-exp}).}
    \label{figure:performance_versus}
\end{figure*}

\section{Experiment settings}

In this section, we describe the models, evaluation benchmarks, and hyperparameters used in our experiments\footnote{Please refer to Appendix~\ref{app-exp} for a detailed description.}.


\paragraph{Vision-Language Models.} To verify the generalization ability of model merging across different model architectures and scales, we conduct experiments on 5 representative MLLMs: \textbf{LLaVA-v1.5-7B (LLaMA-7B)}~\cite{liu2024improvedbaselinesvisualinstruction}, \textbf{Qwen2.5-VL-3B-Instruct (Qwen2.5-3B)}~\cite{qwen2.5,qwen2.5-VL},  \textbf{Qwen2.5-VL-7B-Instruct (Qwen2.5-7B-Instruct)}, \textbf{Qwen3-VL-4B-Instruct (Qwen3-4B)}~\cite{qwen3technicalreport} and \textbf{ Qwen3-VL-8B-Instruct (Qwen3-8B)}. These models span a range of scales from 3B to 8B parameters and encompass both LLaMA-based and Qwen-based architectures, enabling us to assess the robustness of our findings across diverse model families.

\paragraph{Evaluation Benchmarks.} 

We evaluate model performance across a comprehensive suite of vision-language benchmarks that assess diverse multimodal capabilities: \textbf{MMStar}~\cite{chen2024we}, \textbf{MMMU}~\cite{yue2024mmmu}, \textbf{MME}~\cite{fu2025mme}, \textbf{MMBench-EN/CN}~\cite{liu2024mmbench}, \textbf{GQA}~\cite{hudson2019gqa}, \textbf{RealWorldQA}, \textbf{SEED-Bench-2-Plus}~\cite{li2024seed}, and \textbf{POPE}~\cite{li2023evaluating}. These benchmarks assess reasoning, perception, multilingual understanding, visual grounding, and hallucination.

\paragraph{Hyperparameters.}
Throughout our analysis and experiments, we employ a simple merging strategy for all models. We adopt a unified hyperparameter configuration across all models to ensure fair comparison. The merging coefficient $\lambda_i$ is determined through grid search with a step size of 0.1, ranging from 0.0 to 1.0. For the target layer set $\mathcal{L}$, we systematically evaluate different starting points $l_m$ based on the masking analysis in~\Cref{lab:analysis}, typically selecting the transition point where vision information transitions from beneficial to redundant. Unless otherwise specified, we report results using the optimal hyperparameters identified for each model-task combination.

\begin{table*}[!h]
	\begin{adjustbox}{max width=\textwidth, center}
		\includegraphics[width=\textwidth]{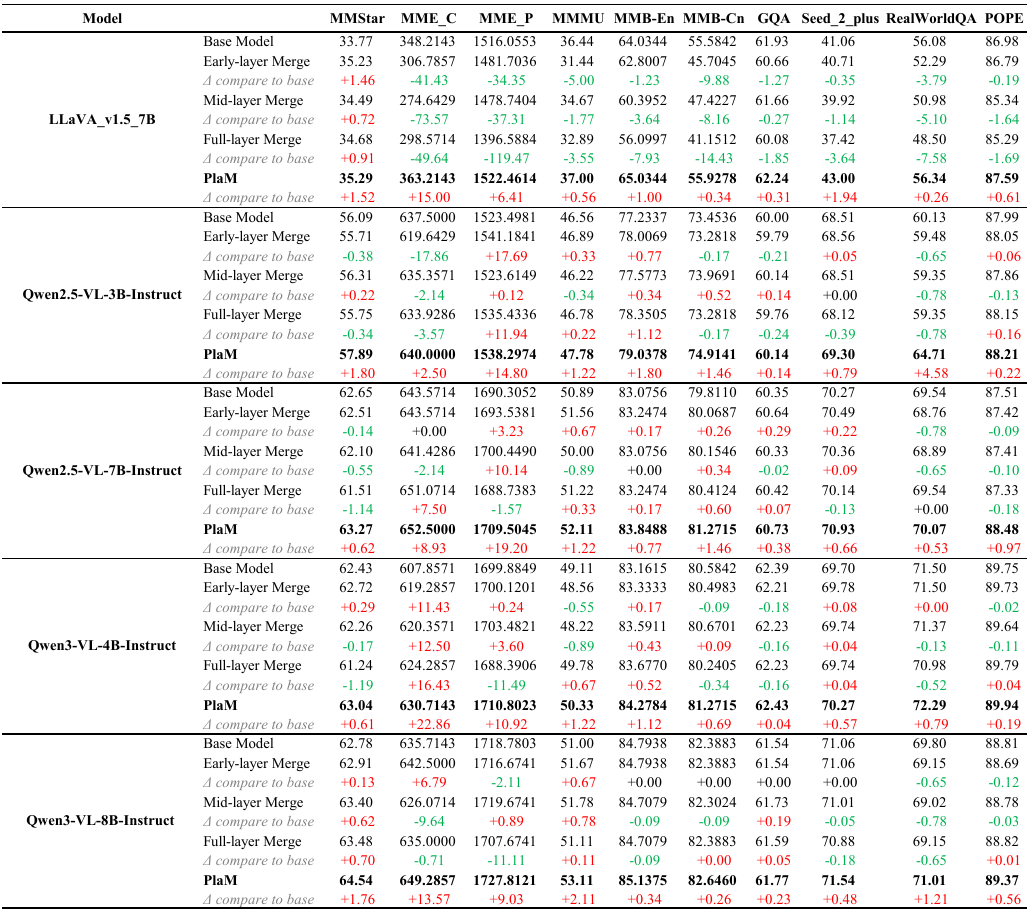}
	\end{adjustbox}
	\caption{Performance comparison of plateau-guided late-layer merging (PlaM) with alternative merging strategies (early-, mid-, and full-layer) across 5 MLLM backbones and 9 benchmarks. We report absolute scores and the change relative to the original fine-tuned MLLM (``Base Model'') for each strategy. PlaM consistently achieves the best overall performance.}
    \label{tab:one}
\end{table*}

\begin{figure*}[t]
    \centering
    \includegraphics[width=1.0\linewidth]{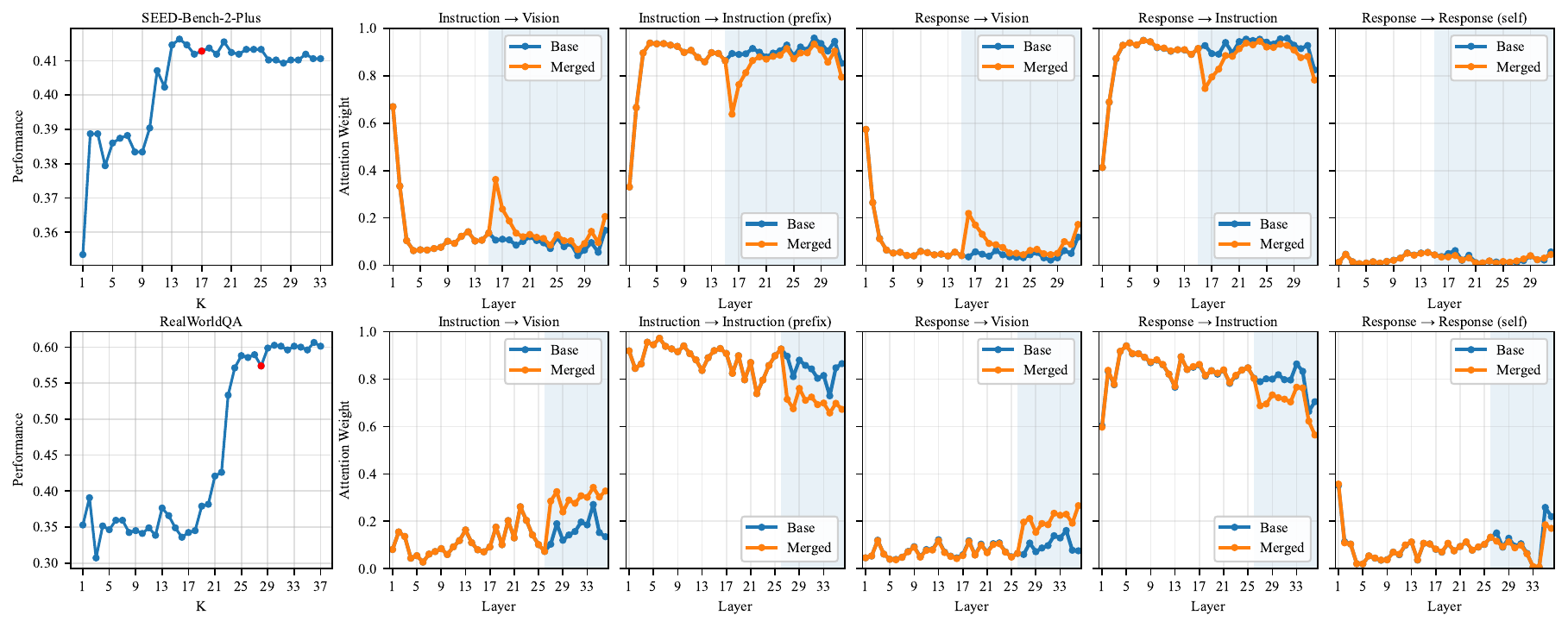}
    \caption{Overall comparison of performance and attention mass before and after merging.
\textbf{Left:} Layer-wise vision token masking results, where the model performance is measured by progressively removing vision tokens from layer $k$ to the last layer. The red marker indicates the selected merge start layer $k_0$, beyond which visual information yields diminishing performance gains and merging is applied. \textbf{Right:} Layer-wise attention mass profiles comparing the original model (Base) and the merged model (Merged). The shaded region denotes the merged layers $\{k_0,L\}$. Results are shown for LLaVA-v1.5-7B on SEED-Bench-2-Plus (top row) and Qwen2.5-VL-3B-Instruct on RealWorldQA (bottom row).}
    \label{fig:mass}
\end{figure*}

\section{Visual Degeneration}

\subsection{Visual Information Absorption}\label{lab:analysis}
Following~\Cref{lab:Preliminary}, we quantify the contribution of visual tokens at different decoder depths by sweeping the cut layer $k$ in the depth-controlled masking intervention. Specifically, for each model and each benchmark, we evaluate the model under a series of settings $k \in \{1,...,L+1\}$. When $k=L+1$, no masking is applied (the original model).
Figure~\ref{figure:performance_versus} illustrates the trend of model performance as $k$ varies.

We observe that different MLLMs display a consistent three-stage trend across various tasks. Based on the points of abrupt change in model performance, the model's behavior can be divided into three stages: (i) early: feature separation; (ii) middle: feature alignment and fusion; (iii) late: feature degradation and absorption.
The input layer of the MLLMs receives information from different modalities and gradually aligns and fuses the information. In the early stages (low $k$), the model primarily focuses on intramodal modeling, during which it is still learning visual and textual semantic representations. The textual representations largely rely on the prior knowledge of the LLM, while cross-modal interactions remain shallow. Consequently, the alignment between visual information and linguistic concepts is weak, and although the incorporation of visual information is beneficial, the resulting performance gains are limited.

As models transition into the middle stage (middle $k$), a pronounced and rapid performance increase is observed across almost all benchmarks, including MMMU, GQA, and MMBench in both English and Chinese. This stage corresponds to the effective alignment of visual and language representations in a shared semantic space. Visual objects, attributes, and spatial relations become consistently mapped to linguistic tokens, enabling the model to move beyond surface-level description toward image-grounded inference. Notably, model scale plays a critical role in this phase: larger models such as Qwen3-VL-8B and Qwen2.5-VL-7B not only achieve higher performance but also maintain a wider and more stable alignment region. This suggests that the middle stage constitutes the core window in which multimodal capabilities are formed and expressed.

In the later stages (high $k$), model performance gradually reaches a plateau, in some cases, may even slightly decline. A plausible explanation is that deeper multimodal fusion increasingly favors abstract linguistic representations, with visual information being progressively compressed or absorbed into the linguistic space. Consequently, in the later layers, performance does not significantly degrade even when visual tokens receive reduced attention. While such deep fusion benefits abstract reasoning, it can also erode the language model’s inherent distributed knowledge, thereby disrupting the robust reasoning patterns acquired during pre-training~\cite{amariucai2023acquiring,zhai2023investigating}. From a representation learning perspective, this suggests that visual signals have already been sufficiently integrated by the later stage, and that \textbf{further performance bottlenecks stem from the degradation of linguistic representations rather than from limitations in visual modeling}.



\subsection{Plateau-Guided Late-Layer Merging}
\label{sec:plateau_merge}
Motivated by the plateau behavior observed in \Cref{lab:analysis}, we propose PlaM that injects base language model parameters into late decoder layers starting from the plateau onset. 
Intuitively, since these layers contribute little additional benefit from continued visual access, we use them to recover general-purpose language model capabilities while preserving multimodal alignment in earlier layers.


For each model, we identify the plateau onset layer $k^\star$ from the performance--$k$ curves (Fig.~\ref{figure:performance_versus}) as the elbow point where the rapid-gain regime transitions into a stable plateau. 
The merge start layer is treated as a lightweight hyperparameter $k_0$, whose value is
determined via a nearest-neighbor search in the vicinity of $k^\star$. All decoder layers from $k_0$ to the final layer are then merged, i.e.,
$\mathcal{L}(k_0) = \{k_0, \dots, L\}$.
Considering that the attention mechanism fundamentally governs how text tokens attend to and integrate visual features in the Transformer decoder~\cite{wu2024controlmllm,kangsee}, 
we merge only the self-attention projections (Q/K/V/O) of the fine-tuned MLLM with the corresponding projections from the base language model described in~\Cref{lab:Preliminary}.

\paragraph{Results.}

\textbf{Overall Results and Comparison with Merging Baselines.} Table~\ref{tab:one} reports the performance comparison between our plateau-guided late-layer merging (PlaM) and three alternative merging strategies: early-layer merging, mid-layer merging, and full-layer merging across all decoder layers. PlaM consistently outperforms the base model across all 5 evaluated architectures and 9 benchmarks, demonstrating its effectiveness as a training-free enhancement method. For example, on LLaVA-v1.5-7B, PlaM improves over the base MLLM on MMStar (35.29 vs. 33.77), MMMU (37.00 vs. 36.44), and RealWorldQA (56.34 vs. 56.08), while maintaining or improving performance on most other benchmarks. Similar trends are observed across the Qwen model families. In contrast, the three alternative merging strategies (Early-layer, Mid-layer and Full-layer) fail to achieve the best overall performance. 
These observations suggest that the effectiveness of PlaM critically depends on where the merging is applied. 
The early and middle layers are primarily responsible for intra-modal feature learning and cross-modal alignment. 
In the early and middle stages, multimodal alignment remains incomplete. Performing parameter merging at this stage causes the model to become overly dependent on the pretrained LLM, thereby inhibiting necessary visual adaptation, disrupting cross-modal alignment, and ultimately degrading overall model performance.
In the late stage, as discussed above, performance bottlenecks stem from the degradation of linguistic representations rather than from limitations in visual modeling.
Merging the original LLM weights at this stage effectively restores part of the text-only representational manifold that may have been distorted during multimodal fine-tuning. By reintroducing the original weights, the model regains access to high-quality linguistic abstractions and stable reasoning trajectories, which can compensate for the noise and over-compression introduced by aggressive cross-modal fusion. Importantly, this does not eliminate multimodal capability; instead, it rebalances the contribution of visual and linguistic representations.


\textbf{Task-type analysis.} The gains of PlaM exhibit a clear task-dependent pattern in Table~\ref{tab:one}. PlaM delivers the most pronounced improvements on MMStar, MME, and MMMU, where success strongly depends on late-stage semantic decision making and visual evidence integration. Concretely, PlaM consistently boosts MMStar by $+0.61\sim +1.80$, and yields especially large gains on MME—MME\_C: $+2.50\sim+22.86$ and MME\_P: $+6.41\sim+19.20$ across all five backbones—together with consistent improvements on MMMU ($+0.56\sim+2.11$). This aligns with the design of PlaM: by merging only plateau-phase late-layer attention projections, it mainly repairs the degraded semantic decision structure that governs which visual cues are retrieved and trusted at the final reasoning stage, thus benefiting benchmarks that are most sensitive to decision-level grounding.
In contrast, PlaM yields moderate but consistent gains on the remaining benchmarks, where performance is often limited by strong baselines and early-to-mid fusion/representation quality rather than late-stage decisions, making late-layer merging less impactful.
Finally, gains are minimal on GQA ($+0.04\sim+0.38$) and POPE ($+0.19\sim+0.97$), where the dominant limitations likely lie beyond late-layer semantic decisions: GQA relies heavily on early-to-mid compositional grounding and spatial/relational modeling, while POPE is closer to evidence-verification and calibration with already high baseline accuracy. Consequently, restoring late-layer semantic decision structure alone provides only marginal benefits on these benchmarks.

\begin{figure*}[t]
    \centering
    \includegraphics[width=1.0\linewidth]{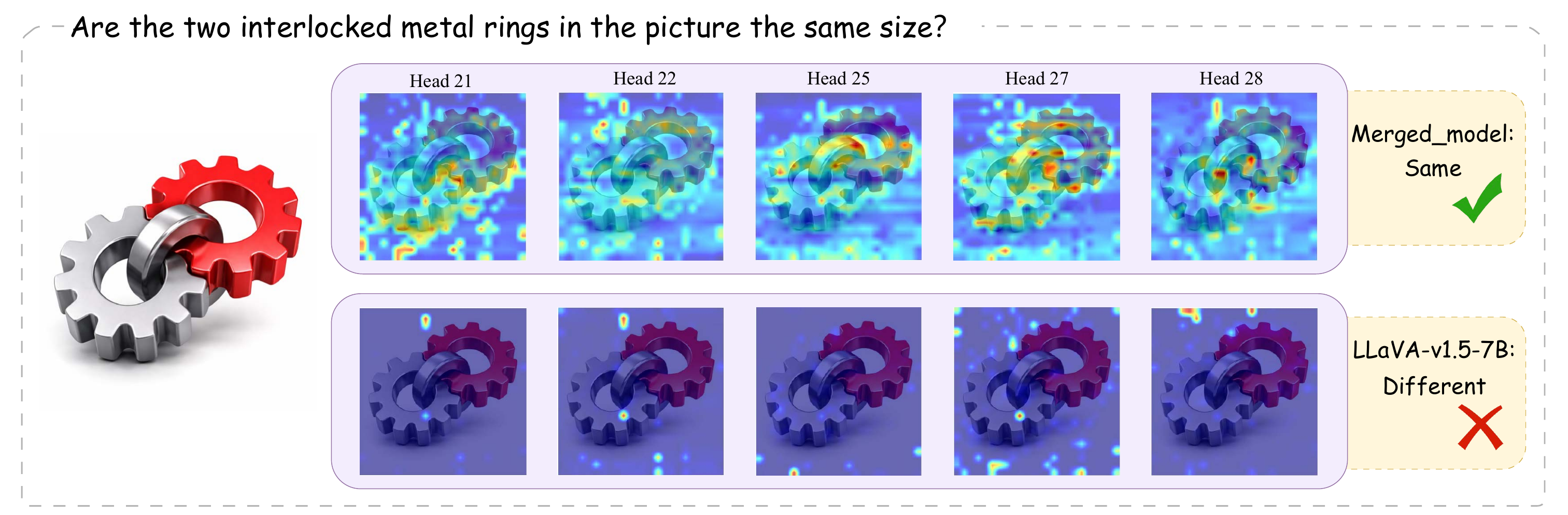}
    \caption{Attention heatmaps for LLaVA-v1.5-7B (bottom) and its PlaM-merged model (top). Additional case studies for other backbones are provided in Appendix~\Cref{figure:heapmap_llava,figure:heapmap_qwen253,figure:heapmap_qwen257,figure:heapmap_qwen34,figure:heapmap_qwen38}.}
    \label{figure:heapmap}
\end{figure*}



\section{Attention-based Mechanistic Analysis}

\paragraph{Measuring Vision Token Attention} We partition the input sequence by token type as $X=[T_\text{pre};T_\text{vis};T_\text{ins}]$, where $T_\text{pre}$ denotes prefix text tokens (e.g., system prompts), $T_\text{vis}$ denotes vision tokens and $T_\text{ins}$ denotes instruction text tokens. During generation, the single-step output token is denoted as $T_\text{res}$. At layer $l$, let $\alpha_{ij}^{(l)}$ denote the attention weight from query token $i$ to key token $j$, averaged over all attention heads. For a target token set $T_\text{tgt}$ and a source token set $T_\text{src}$, we define the attention mass as:
\[
\mathrm{Mass}^{(l)}(T_{\text{tgt}}\rightarrow T_{\text{src}})=\frac{1}{|T_{\text{tgt}}|}\sum_{i\in T_{\text{tgt}}}\sum_{j\in T_{\text{src}}} \alpha _{ij}^{(l)} .
\]
We compute this quantity in the prefill stage with $T_{\text{tgt}}=T_{\text{ins}}$, and in the decoding stage with $T_{\text{tgt}}=T_{\text{res}}$. The source set is chosen from $T_{\text{src}} \in \{T_{\text{vis}}, T_{\text{pre}} \cup T_{\text{ins}} , T_{\text{res}}  \}$, yielding layer-wise curves that serve as an attention-based indicator of how different token types are emphasized across layers. Intuitively, a large $\mathrm{Mass}^{(l)}(\cdot\!\rightarrow\!T_{\text{vis}})$ indicates stronger reliance on visual tokens at depth $l$.



\paragraph{Layer-wise Attention Comparison} Figure~\ref{fig:mass} compares the layer-wise attention mass profiles between the original models and their merged counterparts. A consistent pattern emerges across both architectures: merging substantially increases the attention mass allocated to vision tokens. We attribute this shift to the injection of base language model capacity into the late layers, which restores stronger semantic decision-making and, in turn, encourages more active retrieval and utilization of visual evidence.


For Qwen2.5-VL-3B-Instruct (merged layers 27-36), the merged model exhibits a clear upward trend in instruction-to-vision attention mass ($\text{Mass}^{(l)}(T_{\text{ins}}\rightarrow T_{\text{vis}})$) in late layers, rising from near 0.1 to 0.3, which suggests that merging strengthens visual evidence integration during prompt encoding itself.
Meanwhile, the response-to-vision attention mass ($\text{Mass}^{(l)}(T_{\text{res}}\rightarrow T_{\text{vis}})$) shows a modest increase in attention mass toward vision tokens, which further suggests that the merged model performs explicit visual verification or retrieval during output generation and ensures response accuracy and grounding. Similar late-layer increases in vision-directed attention are also observed for LLaVA.
 

It is important to note that higher vision-token attention mass reflects increased reliance on visual information rather than a guarantee of improved reasoning quality. However, the consistent shift toward greater utilization of vision tokens across both architectures aligns well with the performance improvements reported in~\Cref{sec:plateau_merge}. 
This mechanistic evidence supports our hypothesis that plateau-guided merging guides late layers to explore broader visual evidence by recovering the textual capabilities, thereby improving multimodal grounding.

\paragraph{Case Study} 

To further investigate whether the increased attention mass corresponds to more effective visual grounding, we visualize attention heatmaps from several attention heads for LLaVA-v1.5-7B and its PlaM counterpart. Figure~\ref{figure:heapmap} presents the attention patterns for the question "Are the two interlocked metal rings in the picture the same size?", which requires fine-grained visual comparison of the two rings. The contrast is striking: the original LLaVA model (bottom row) exhibits predominantly weak and diffuse attention, with most heads failing to consistently focus on the ring boundaries, and consequently produces an incorrect answer ("Different"). In contrast, the merged model (top row) demonstrates more structured and task-relevant attention patterns. Multiple heads allocate concentrated attention to both rings, particularly along their contours and the interlocking region, thereby enabling accurate size comparison and the correct answer ("Same"). This example suggests that the benefits of PlaM extend beyond higher aggregate attention to vision tokens, manifesting as enhanced spatial localization on decision-critical visual evidence during decoding.

\section{Conclusion}
In this paper, we propose \textbf{PlaM}, a training-free plateau-guided late-layer merging method that injects base LM attention projections into plateau-phase layers to restore degraded semantic decision structure while preserving earlier multimodal alignment. Extensive experiments across five backbones and nine benchmarks show consistent improvements over the original models and alternative merging schemes.
Analyses of attention mass and heatmaps further indicate that PlaM makes late-layer visual attention more focused on semantically relevant regions, helping mitigate scattered attention and improving visual grounding. 
We hope our findings provide actionable insights into layer-wise multimodal behavior and inspire future work on training-free interventions in MLLMs.

\section*{Limitations}
This work has the following limitations.
We introduce a hyperparameter $k_0$, which determines the merging position of the model. The optimal $k_0$ configuration varies for different datasets, and determining the optimal $k_0$ configuration requires more experiments and costs.
We use a simple model merging strategy, more complex model merging methods may yield better results, although this does not conflict with our work.

\bibliography{custom}
\clearpage
\appendix

\section{Experiment settings}\label{app-exp}


\paragraph{Evaluation Benchmarks.}
To comprehensively assess the effectiveness of PlaM, we evaluate performance across a diverse suite of vision-language benchmarks that collectively measure distinct aspects of multimodal understanding and reasoning:
\begin{itemize}[noitemsep, left=0pt]
\item \textbf{MMStar}~\cite{chen2024we}: An elite vision-indispensable benchmark comprising 1{,}500 human-curated samples, designed to ensure strong visual dependency and minimal data leakage, and to evaluate six core capabilities spanning perception, reasoning, and domain knowledge.
\item \textbf{MMMU}~\cite{yue2024mmmu}: A college-level multi-discipline multimodal benchmark that requires expert knowledge across six academic domains, evaluating models' expert-level multimodal understanding.
\item \textbf{MME}~\cite{fu2025mme}: A comprehensive evaluation benchmark assessing both perception and cognition abilities through 14 subtasks.
\item \textbf{MMBench-EN/CN}~\cite{liu2024mmbench}: A bilingual benchmark with objective questions in English and Chinese, assessing perception and reasoning abilities across languages and cultural contexts to assess multilingual multimodal abilities.
\item \textbf{GQA}~\cite{hudson2019gqa}: A visual reasoning benchmark testing compositional question answering and spatial reasoning over real-world images, requiring models to understand complex relationships and perform multi-hop reasoning.
\item \textbf{RealWorldQA}: A real-world spatial understanding benchmark released alongside Grok-1.5 Vision, focusing on real-world images captured from vehicles and other real settings to evaluate practical visual reasoning in real environments.
\item \textbf{SEED-Bench-2-Plus}~\cite{li2024seed}: A text-rich visual comprehension benchmark that evaluates MLLMs’ ability to interpret embedded texts, understand visual content, and model their interactions, consisting of 2.3K human-annotated multiple-choice questions spanning three real-world categories (Charts, Maps, and Webs) with 63 fine-grained types. 
\item \textbf{POPE}~\cite{li2023evaluating}: A polling-based yes/no benchmark for object hallucination that probes object existence with random/popular/adversarial negatives and reports standard classification metrics.
\end{itemize}
This carefully selected benchmark suite provides holistic evaluation spanning complementary dimensions: reasoning depth (from perception to cognition), multilingual understanding, spatial and compositional reasoning, real-world applicability, and reliability (hallucination detection). To perform the evaluation, we use the lmms\_eval library~\cite{zhang2025lmms}. For each dataset, we keep the evaluation setup fixed and apply it consistently across all models.

\paragraph{Hyperparameters.}

\begin{table*}[t]
\centering
\small
\setlength{\tabcolsep}{3pt}        
\renewcommand{\arraystretch}{0.95}
\resizebox{\textwidth}{!}{%
\begin{tabular}{clccccccccc}
\hline
 &  & MMStar & MME & MMMU & MMB-En & MMB-Cn & Seed\_2\_plus & POPE & GQA & RealWorldQA \\ \hline
\multirow{3}{*}{\begin{tabular}[c]{@{}c@{}}LLaVA\_v1.5\_7B\\ (32)\end{tabular}}
 & $\lambda_1$ & 0.6 & 0.1 & 0.5 & 0.3 & 0.3 & 0.2 & 0.1 & 0.2 & 0.1 \\
 & $\lambda_2$ & 0.4 & 0.9 & 0.6 & 0.8 & 0.9 & 0.8 & 1.3 & 1.1 & 0.9 \\
 & $k_0$       & 20  & 18  & 20  & 22  & 25  & 16  & 19  & 29  & 29  \\ \hline

\multirow{3}{*}{\begin{tabular}[c]{@{}c@{}}Qwen2.5-VL-3B-Instruct\\ (36)\end{tabular}}
 & $\lambda_1$ & 0.4 & 0.1 & 0.7 & 0.4 & 0.3 & 0.3 & 0.1 & 0.1 & 0.2 \\
 & $\lambda_2$ & 0.7 & 0.9 & 0.4 & 0.6 & 0.8 & 0.7 & 0.9 & 0.9 & 0.7 \\
 & $k_0$       & 22  & 22  & 24  & 24  & 21  & 28  & 26  & 25  & 27  \\ \hline

\multirow{3}{*}{\begin{tabular}[c]{@{}c@{}}Qwen2.5-VL-7B-Instruct\\ (28)\end{tabular}}
 & $\lambda_1$ & 0.1 & 0.1 & 0.4 & 0.3 & 0.5 & 0.3 & 0.9 & 0.2 & 0.4 \\
 & $\lambda_2$ & 0.9 & 0.9 & 0.8 & 0.8 & 0.6 & 1.0 & 0.2 & 0.8 & 0.7 \\
 & $k_0$       & 19  & 22  & 22  & 18  & 20  & 22  & 20  & 18  & 24  \\ \hline

\multirow{3}{*}{\begin{tabular}[c]{@{}c@{}}Qwen3-VL-4B-Instruct\\ (36)\end{tabular}}
 & $\lambda_1$ & 0.3 & 0.2 & 0.1 & 0.2 & 0.3 & 0.1 & 0.3 & 0.1 & 0.4 \\
 & $\lambda_2$ & 0.9 & 0.9 & 1.2 & 1.0 & 0.9 & 1.0 & 0.8 & 1.0 & 0.4 \\
 & $k_0$       & 22  & 22  & 25  & 21  & 21  & 19  & 19  & 29  & 28  \\ \hline

\multirow{3}{*}{\begin{tabular}[c]{@{}c@{}}Qwen3-VL-8B-Instruct\\ (36)\end{tabular}}
 & $\lambda_1$ & 0.3 & 0.1 & 0.3 & 0.2 & 0.3 & 0.3 & 0.2 & 0.1 & 0.4 \\
 & $\lambda_2$ & 0.9 & 1.0 & 0.8 & 0.7 & 0.7 & 0.9 & 0.8 & 0.9 & 0.6 \\
 & $k_0$       & 25  & 20  & 19  & 21  & 25  & 21  & 20  & 18  & 18  \\ \hline
\end{tabular}%
}
\caption{Best hyperparameter settings for PlaM across five MLLM backbones and nine benchmarks.}
\end{table*}

\begin{figure*}[!t]
    \centering
    \includegraphics[width=1.0\linewidth]{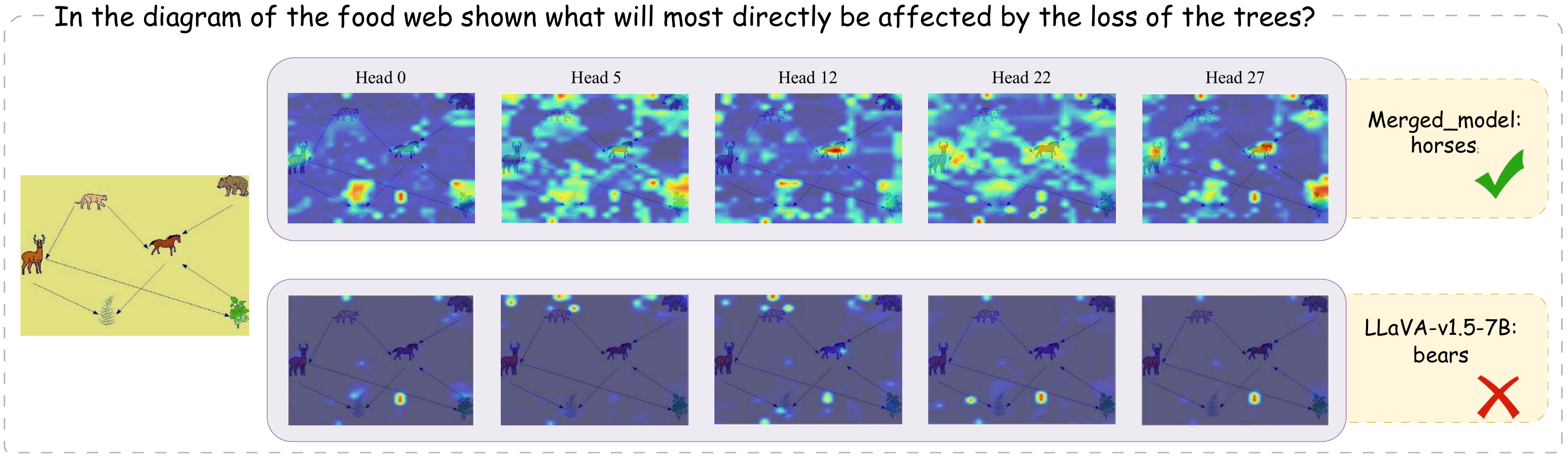}
    \caption{Attention heatmaps for LLaVA-v1.5-7B (bottom) and its PlaM-merged model (top).}
    \label{figure:heapmap_llava}
\end{figure*}

\begin{figure*}[!t]
    \centering
    \includegraphics[width=1.0\linewidth]{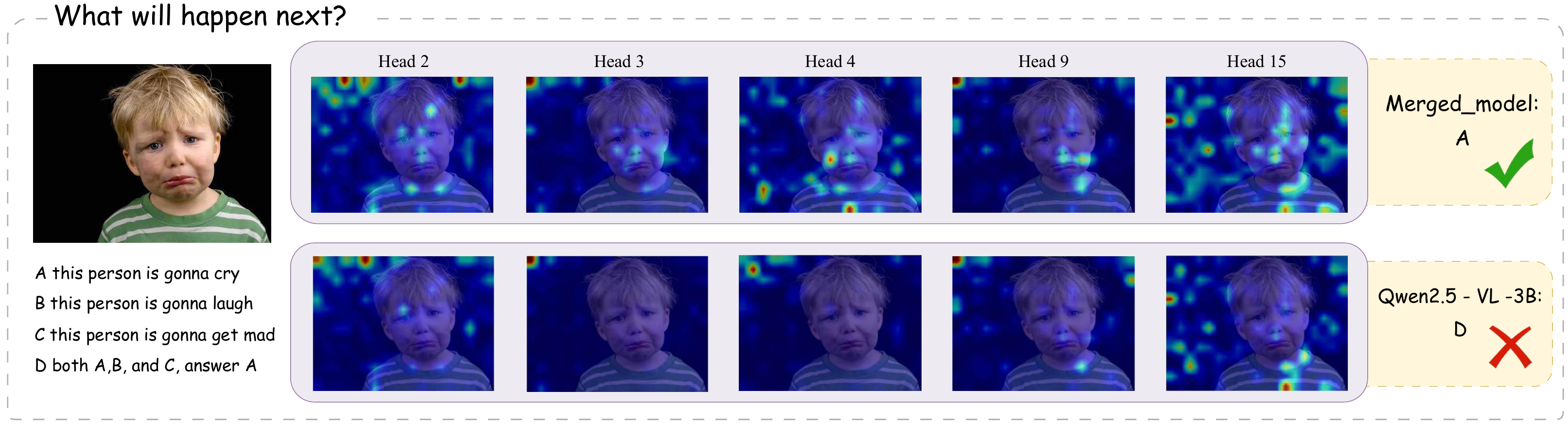}
    \caption{Attention heatmaps for Qwen2.5-VL-3B-Instruct (bottom) and its PlaM-merged model (top).}
    \label{figure:heapmap_qwen253}
\end{figure*}

\begin{figure*}[!t]
    \centering
    \includegraphics[width=1.0\linewidth]{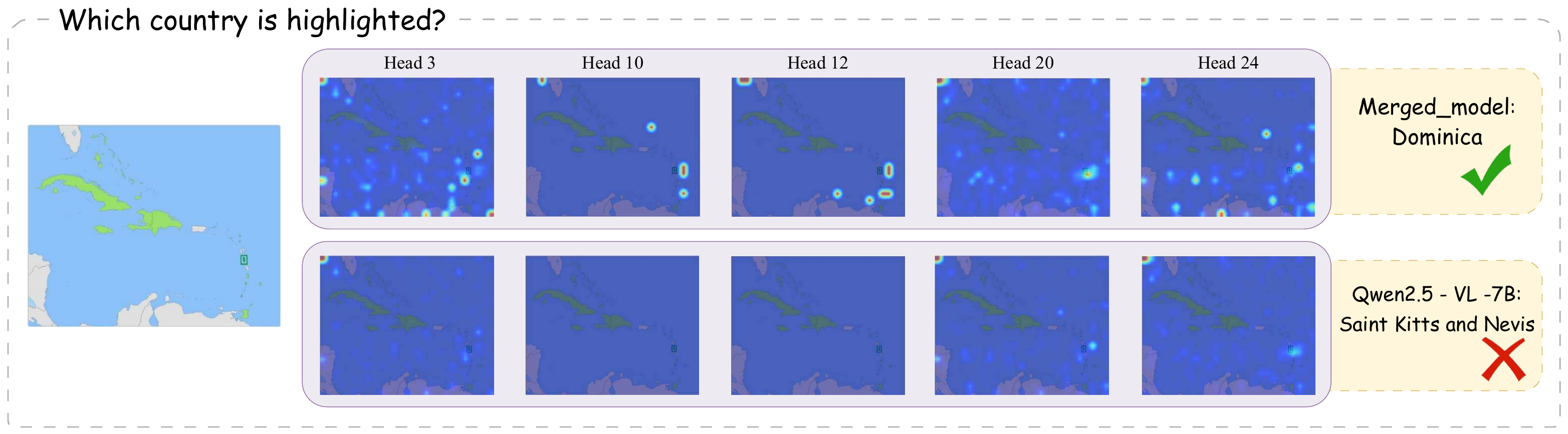}
    \caption{Attention heatmaps for Qwen2.5-VL-7B-Instruct (bottom) and its PlaM-merged model (top).}
    \label{figure:heapmap_qwen257}
\end{figure*}

\begin{figure*}[!t]
    \centering
    \includegraphics[width=1.0\linewidth]{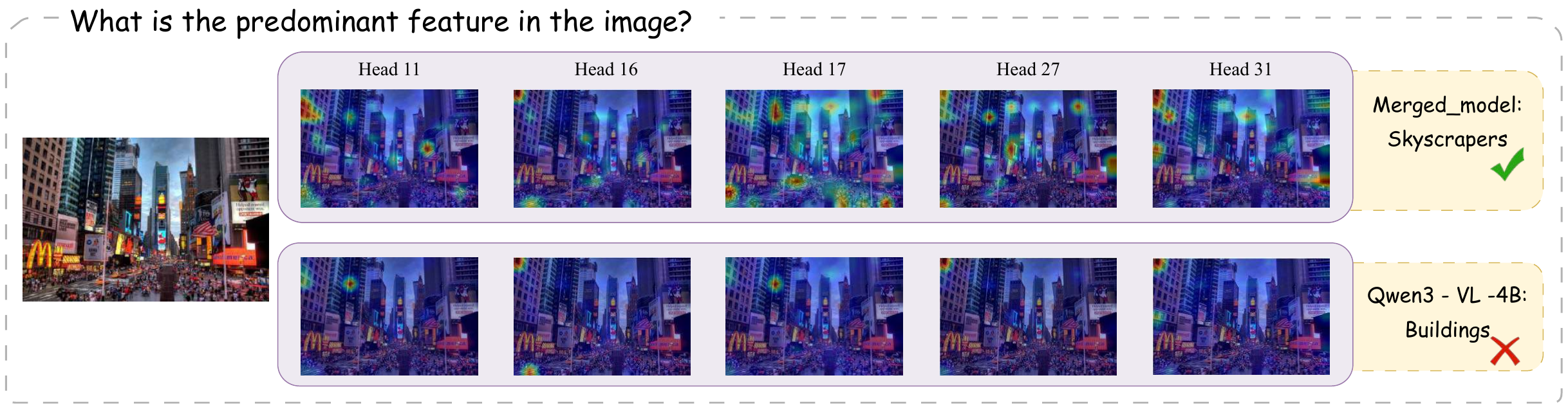}
    \caption{Attention heatmaps for Qwen3-VL-4B-Instruct (bottom) and its PlaM-merged model (top).}
    \label{figure:heapmap_qwen34}
\end{figure*}

\begin{figure*}[!t]
    \centering
    \includegraphics[width=1.0\linewidth]{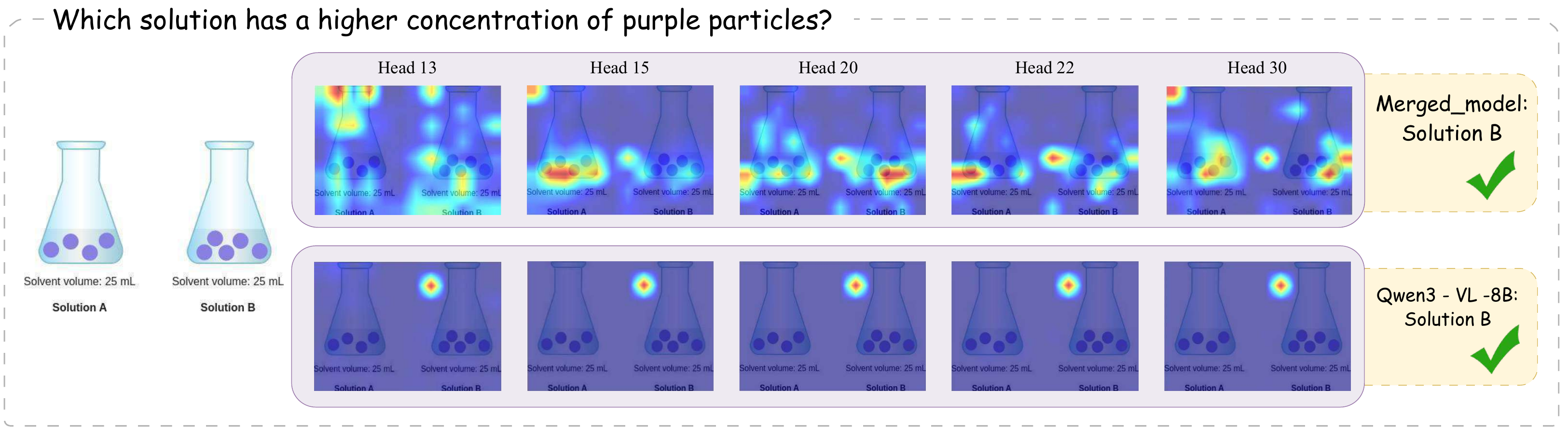}
    \caption{Attention heatmaps for Qwen3-VL-8B-Instruct (bottom) and its PlaM-merged model (top).}
    \label{figure:heapmap_qwen38}
\end{figure*}

\begin{table*}[!h]
	\begin{adjustbox}{max width=\textwidth, center}
		\includegraphics[width=\textwidth]{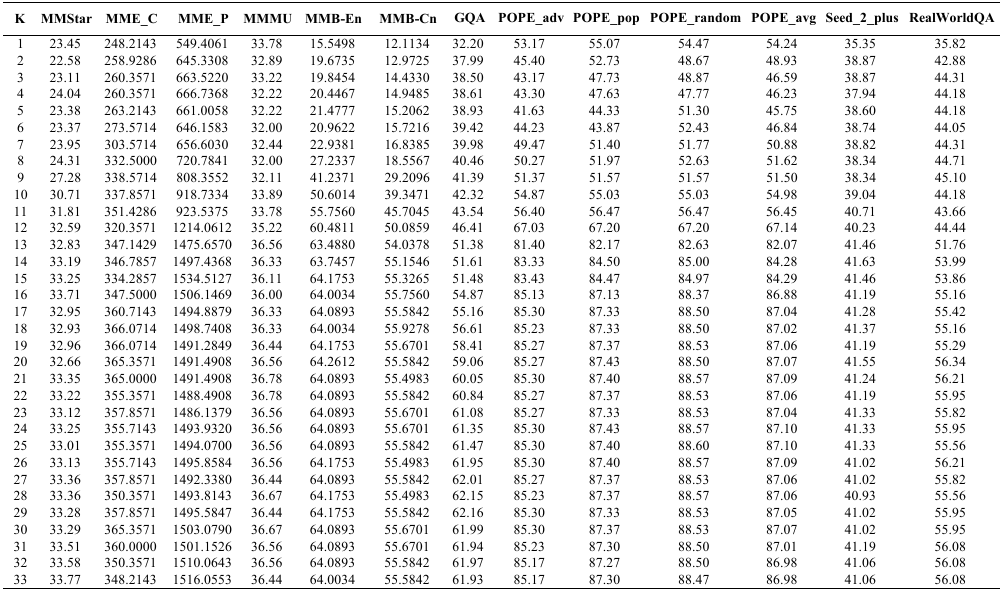}
	\end{adjustbox}
	\caption{Raw scores of LLaVA-v1.5-7B under depth-controlled vision token masking. Scores across benchmarks for each cut layer $k$ (used to generate~\Cref{figure:performance_versus}).}
    \label{tab:llavav1.5}
\end{table*}

\begin{table*}[!h]
	\begin{adjustbox}{max width=\textwidth, center}
		\includegraphics[width=\textwidth]{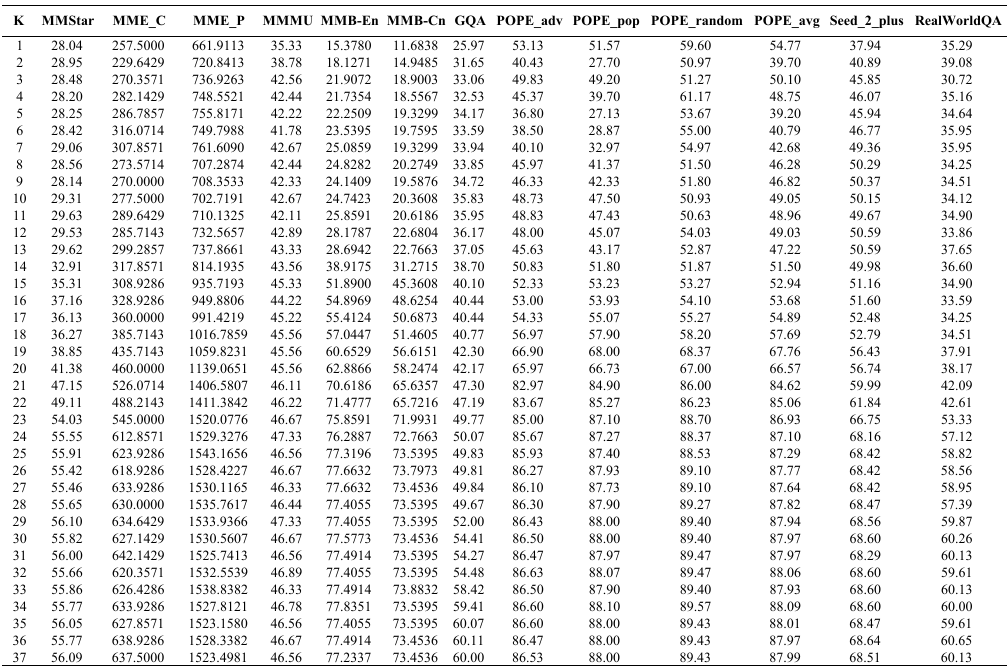}
	\end{adjustbox}
	\caption{Raw scores of Qwen2.5-VL-3B-Instruct under depth-controlled vision token masking. Scores across benchmarks for each cut layer $k$ (used to generate~\Cref{figure:performance_versus}).}
    \label{tab:Qwen2.53B}
\end{table*}

\begin{table*}[!h]
	\begin{adjustbox}{max width=\textwidth, center}
		\includegraphics[width=\textwidth]{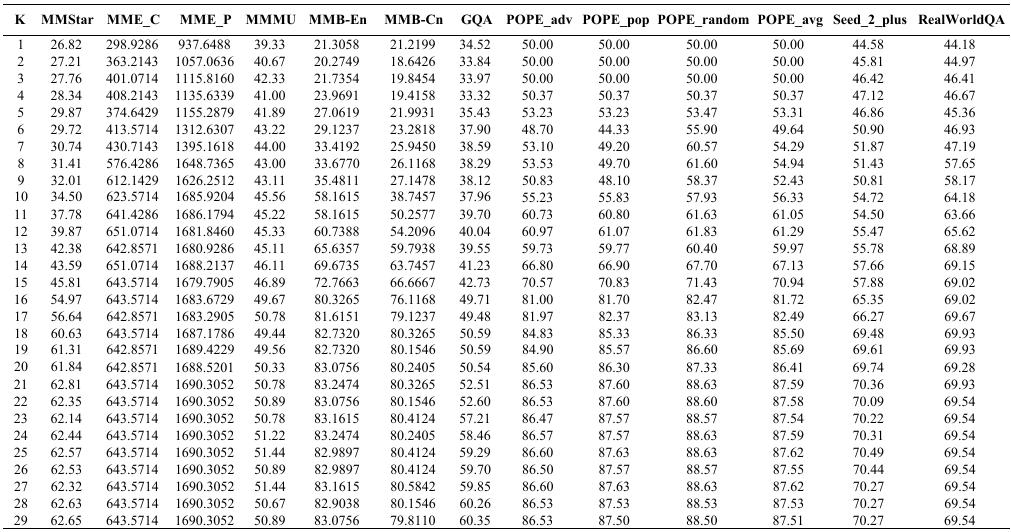}
	\end{adjustbox}
	\caption{Raw scores of Qwen2.5-VL-7B-Instruct under depth-controlled vision token masking. Scores across benchmarks for each cut layer $k$ (used to generate~\Cref{figure:performance_versus}).}
    \label{tab:Qwen2.57B}
\end{table*}

\begin{table*}[!h]
	\begin{adjustbox}{max width=\textwidth, center}
		\includegraphics[width=\textwidth]{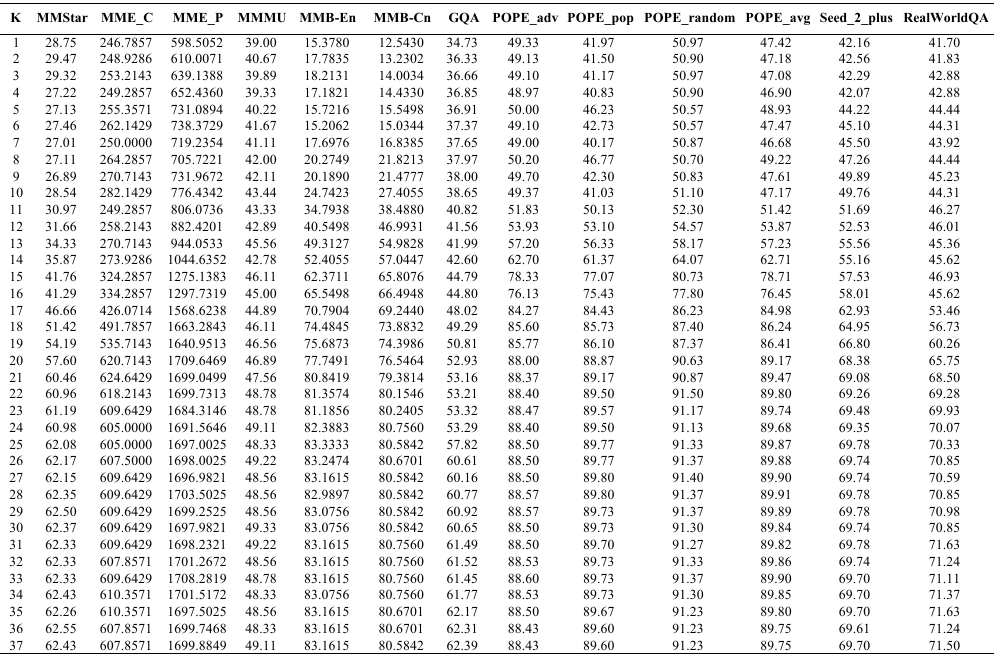}
	\end{adjustbox}
	\caption{Raw scores of Qwen3-VL-4B-Instruct under depth-controlled vision token masking. Scores across benchmarks for each cut layer $k$ (used to generate~\Cref{figure:performance_versus}).}
    \label{tab:Qwen34B}
\end{table*}

\begin{table*}[!h]
	\begin{adjustbox}{max width=\textwidth, center}
		\includegraphics[width=\textwidth]{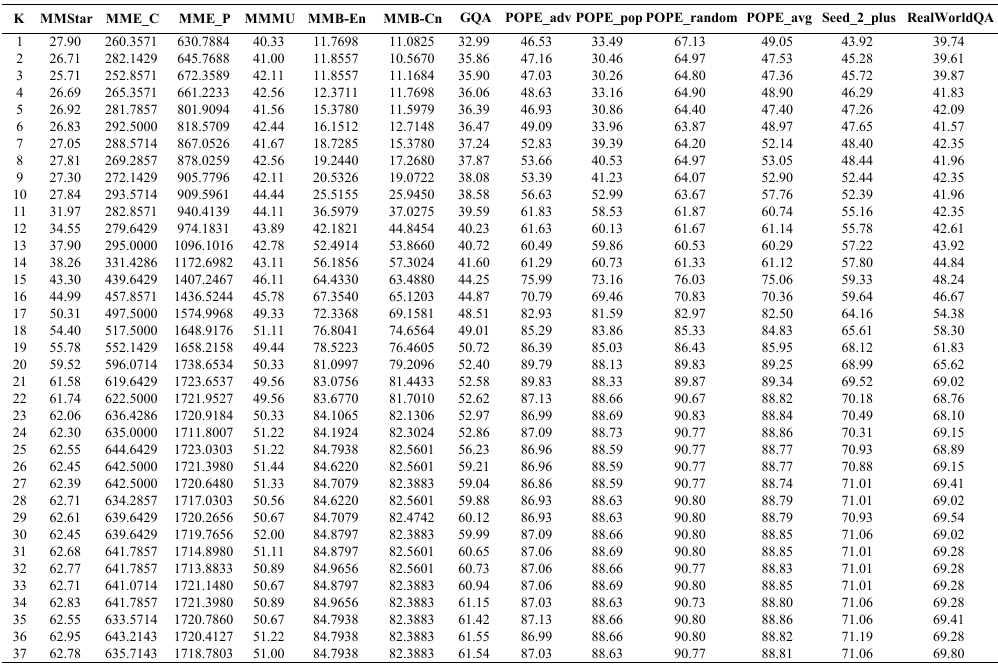}
	\end{adjustbox}
	\caption{Raw scores of Qwen3-VL-8B-Instruct under depth-controlled vision token masking. Scores across benchmarks for each cut layer $k$ (used to generate~\Cref{figure:performance_versus}).}
    \label{tab:Qwen38B}
\end{table*}

\end{document}